\documentclass[iicol,sn-chicago]{sn-jnl}

\usepackage{graphicx}%
\usepackage{multirow}%
\usepackage{amsmath,amssymb,amsfonts}%
\usepackage{amsthm}%
\usepackage{mathrsfs}%
\usepackage[title]{appendix}%
\usepackage{xcolor}%
\usepackage{textcomp}%
\usepackage{manyfoot}%
\usepackage{booktabs}%
\usepackage{algorithm}%
\usepackage{braket}%
\usepackage{algorithmicx}%
\usepackage{algpseudocode}%
\usepackage{listings}%
\usepackage{subcaption}
\usepackage{tablefootnote}
\theoremstyle{thmstyleone}%
\theoremstyle{thmstyletwo}%

\theoremstyle{thmstylethree}%

\begin{document}

\title[Article Title]{Improved Differential Evolution based Feature Selection through Quantum, Chaos, and Lasso}


\author[1,3]{\fnm{Yelleti} \sur{Vivek}}\email{yvivek@idrbt.ac.in}

\author[2]{\fnm{Sri Krishna} \sur{Vadlamani}}\email{srikv@mit.edu}

\author*[1]{\fnm{Vadlamani} \sur{Ravi}}\email{vravi@idrbt.ac.in}

\author[3]{\fnm{P. Radha} \sur{Krishna}}\email{prkrishna@nitw.ac.in}

\affil*[1]{\orgdiv{Centre for Artificial Intelligence and Machine Learning}, \orgname{Institute for Development and Research in Banking Technology}, \orgaddress{\street{Castle Hills Road \#1, Masab Tank}, \city{Hyderabad}, \postcode{500076}, \state{Telangana}, \country{India}}}

\affil[2]{\orgdiv{Research Laboratory of Electronics}, \orgname{Massachusetts Institute of Technology}, \orgaddress{ \postcode{MA}, \state{Cambridge}, \country{United States of America}}}

\affil[3]{\orgdiv{Department of Computer Science and Engineering}, \orgname{National Institute of Technology Warangal}, 
\orgaddress{\city{Warangal}, \postcode{506004}, \state{Telangana}, \country{India}}}


\abstract{Modern deep learning continues to achieve outstanding performance on an astounding variety of high-dimensional tasks. In practice, this is obtained by fitting deep neural models to all the input data with minimal feature engineering, thus sacrificing interpretability in many cases. However, in applications such as medicine, where interpretability is crucial, feature subset selection becomes an important problem. Metaheuristics such as Binary Differential Evolution are a popular approach to feature selection, and the research literature continues to introduce novel ideas, drawn from quantum computing and chaos theory, for instance, to improve them. In this paper, we demonstrate that introducing chaos-generated variables, generated from considerations of the Lyapunov time, in place of random variables in quantum-inspired metaheuristics significantly improves their performance on high-dimensional medical classification tasks and outperforms other approaches. We show that this chaos-induced improvement is a general phenomenon by demonstrating it for multiple varieties of underlying quantum-inspired metaheuristics. Performance is further enhanced through Lasso-assisted feature pruning. At the implementation level, we vastly speed up our algorithms through a scalable island-based computing cluster parallelization technique.}

\keywords{Feature Subset Selection; Differential Evolution; Quantum inspired algorithms; Chaos; Big Data}


\maketitle

\section{Introduction}\label{sec1}
The fundamental objective of feature selection is to identify the most important and discriminative features from a given set of features. Its prominence has garnered the attention of researchers and practitioners in domains replete with big, high-dimensional datasets. Feature selection \citep{ref39,ref40} entails a great reduction in computational complexity, improvement in human comprehensibility, and easy deployment of models in production.
Wrapper methods based on metaheuristics solve the feature subset selection (FSS) problem by posing it as a combinatorial optimization problem. These methods attempt to identify an efficient feature subset with the least cardinality and associated high accuracy among the $2^{n}-1$ possible number of combinations, where $n$ is the total number of features in the dataset. Among the metaheuristics employed for this purpose, evolutionary algorithms have been proven to be most efficient for determining optimal feature subsets owing to the inherent parallelism present in the population-based search \citep{ref43,ref44}. 
\begin{table}[hbtp]
    \centering
    \caption{Notation used in the current study}
    \label{tab:notation}
        \begin{tabular}{|p{2.5cm}|p{4cm}|}
        \hline
        $\phi$ & Empty set \\ \hline
        $N$ & Population size \\ \hline
        $X$ & Population comprising N number of solutions \\ \hline
        $n$ & Number of dimensions / features \\ \hline
        $X_i$ & i\textsuperscript{th} solution of population X and having $n$ dimensions \\ \hline
        $M^t$ & Mutated vectors at generation $t$ comprising $N$ solutions \\ \hline
        $M_i^t$ & $i\textsuperscript{th}$ solution of mutation population of $n$ dimensions \\ \hline
        F & Mutation factor \\ \hline
        $U^t$ & trial vectors at generation t comprising ps solutions \\ \hline
        $U_i^t$ & $i\textsuperscript{th}$ solution of mutation population of $n$ dimensions \\ \hline
        $CR$ & Crossover rate \\ \hline
        $MAXITR$ & Maximum number of iterations \\ \hline
        $randi$ & Randomly chosen index \\ \hline
        $d_t$ & Chaotic number at t\textsuperscript{th} time step \\ \hline
        $rand(0,1)$ & Random number generated between $[0,1]$ \\ \hline
        $\lambda$ & Logistic map control parameter \\ \hline
        AUC$_i$ & AUC score of an i\textsuperscript{th} solution \\ \hline
        cardinality$_i$ & Cardinality score of an $i\textsuperscript{th}$ solution \\ \hline
        $cvalue_t$ & Chaotic random number at $t\textsuperscript{th}$ time step \\ \hline
        $P$ & RDD of population X \\ \hline
        $Xtrain$ & Train dataset \\ \hline
        $Xtest$ & Test dataset \\ \hline
        $localN$ & Local population size \\ \hline
        $mMig$ & Maximum number of migrations \\ \hline
        $mGen$ & Maximum number of generations \\ \hline
        $localN$ & Local population / sub-population pertained to a data island comprising lps solutions \\ \hline
    \end{tabular}
    \label{tab:notation}
\end{table}
Among the evolutionary algorithms, Differential Evolution (DE) proved to be robust while solving many combinatorial and continuous optimization problems \cite{ref0,ref41,ref42}. 
Despite its supremacy over other algorithms, DE suffers from a tendency to get stuck prematurely in local optimal solutions, which affects its exploration and exploitation capabilities \cite{ref0}. To alleviate these issues, researchers proposed several quantum-inspired algorithms (QIEA) to solve a myriad of combinatorial optimization problems such as knapsack problem, truck trail problem, and portfolio optimization. Often, QIEAs accelerate the evolution process owing to their quantum parallelization \citep{ref01,ref1} and entanglement of the quantum state. Further, the diversity and convergence rate are improved well enough to increase the probability of getting better global optimal solutions.

In today's data-rich environment, the humongous growth of high-dimensional datasets warrants the critical need for developing scalable algorithms \citep{ref45,ref46}. Despite the popularity of Hadoop and its ecosystem as a big data framework, Spark's unique features, including in-memory computing and seamless integration, have made it a viable and competitive alternative to Hadoop \citep{ref47}. However, the extant Quantum-Inspired Evolutionary Algorithms (QIEAs) are not scalable to large, high-dimensional datasets.  

Our contributions in this paper include:
\begin{itemize}
    \item proposing chaotic, quantum-inspired evolutionary algorithms for FSS in high-dimensional data and demonstrating their superiority over the extant methods on four problems.
    \item utilizing the Lyapunov exponent to ensure that we work in a truly chaotic regime.
    \item introducing LASSO LR in place of LR as a classifier for the FSS wrapper
    \item proposing parallel versions of the above-mentioned algorithms operating in a single-objective environment under an island-based approach in Apache Spark framework.
\end{itemize}

The paper is structured as follows: Section 2 reviews the literature, Section 3 presents our new algorithms, Section 4 describes the datasets analyzed in the study, Section 5 analyses the results obtained, and section 6 concludes the paper. 

\begin{table*}[htbp]
    \centering
    \caption{Full form of the acronyms of the Algorithms employed in the current study}
    \label{tab:algorithms}
    \begin{tabular}{|p{1cm}|p{2cm}|p{11cm}|}
     \hline
       No. & Algorithm  & Description \\
       \hline
       1 & LR & Logistic Regression \\
       \hline
       2 & LLR & Least Absolute Shrinkage and Selection Operator LR\\
       \hline
       3 & BDE & non-quantum counterpart of QBDE\\
       \hline
       4 & QBDE-I & a non-gate quantum variant of BDE with threshold trick and random numbers generated from the uniform distribution\\
       \hline
       5 & QBDE-II & a gate quantum variant of BDE with threshold trick and random numbers generated from uniform distribution\\
       \hline
       6 & CQBDE-I & a chaotic maps guided variant of QBDE-I but without Lyapunov exponent guidance\\
       \hline
       7 & CQBDE-II & a chaotic variant of QBDE-II without Lyapunov exponent guidance\\
       \hline
       8 & CLQBDE-I & a chaotic variant with Lyapunov exponent of QBDE-I\\
       \hline
       9 & CLQBDE-II & a chaotic variant with Lyapunov exponent of QBDE-II \\
       \hline
       10 & CQIEA & an algorithm from \cite{ref35} \\
       \hline
       11 & CTQIEA & a variant of CQIEA \citep{ref35}which integrates threshold trick and chaos without Lyapunov exponent guidance\\
       \hline
       12 & CLTQIEA &  a variant of CQIEA \citep{ref35} which incorporates both threshold trick and Lyapunov exponent guidance.\\
       \hline
    \end{tabular}
    \label{tab:my_label}
\end{table*} 

\begin{figure*}[hbtp]
\centering
\begin{subfigure}[b]{0.48\textwidth}
    \centering
    \includegraphics[width=1.1\columnwidth]{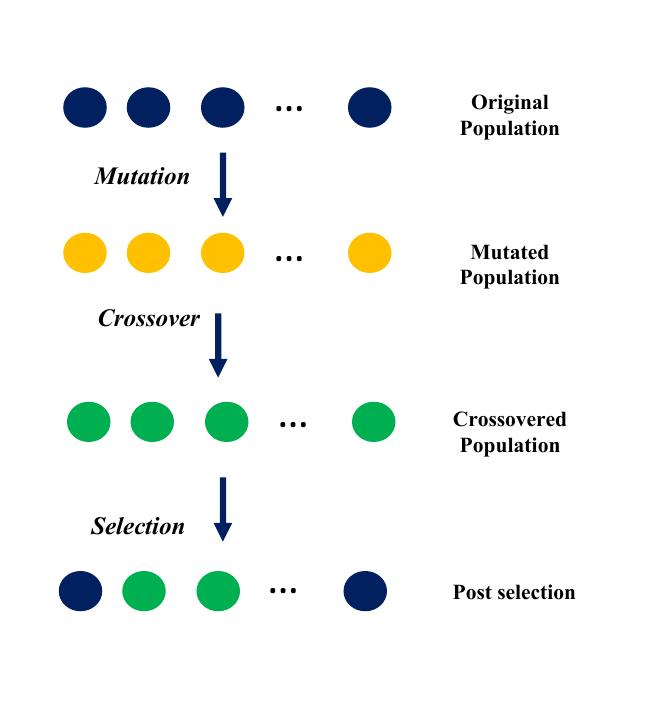}
    \caption{Schematic of the BDE}
    \label{BDE}
\end{subfigure}
\hfill
\begin{subfigure}[b]{0.48\textwidth}
    \centering
    \includegraphics[width=1.1\columnwidth]{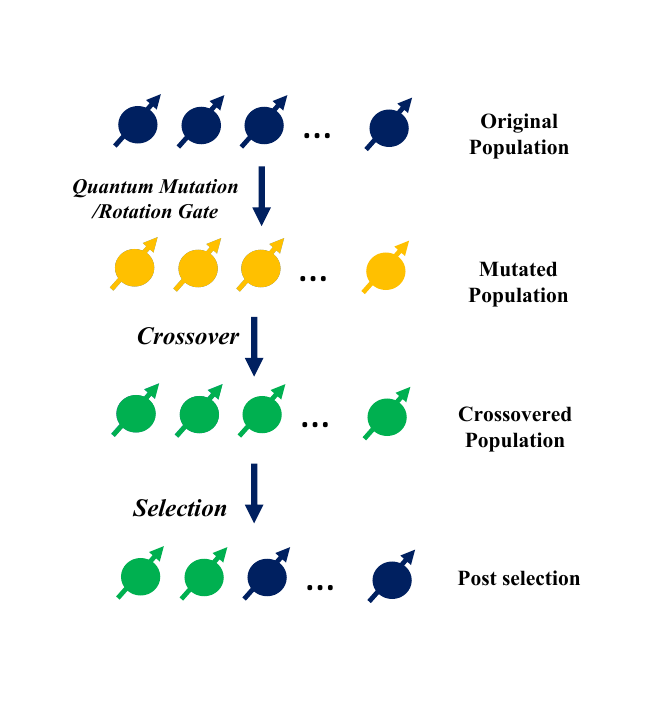}
    \caption{Schematic of the proposed QBDE variants}
    \label{QBDE}
\end{subfigure}
\hfill
\begin{subfigure}[b]{0.52\textwidth}
    \centering
    \includegraphics[width=1.1\columnwidth]{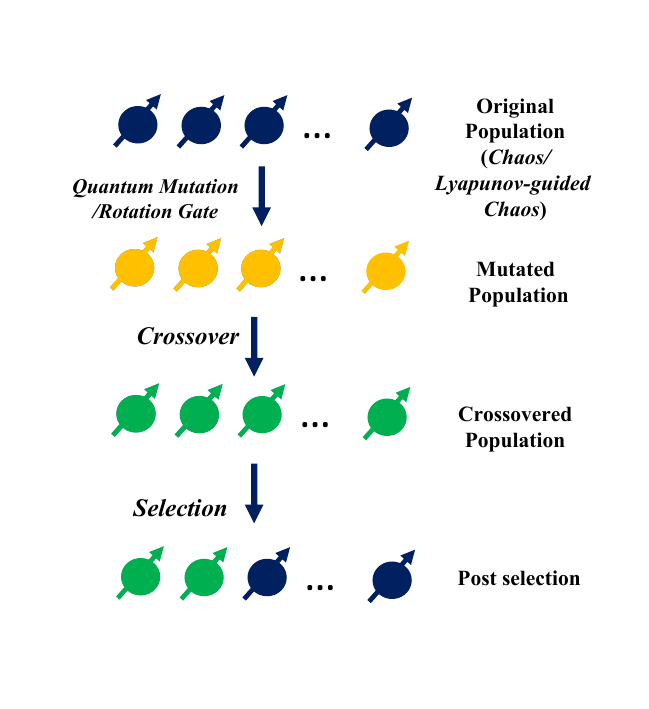}
    \caption{Schematic of the proposed CQBDE variants}
    \label{CQBDE}
\end{subfigure}
\begin{subfigure}[b]{0.48\textwidth}
\centering
\includegraphics[width=1.2\columnwidth]{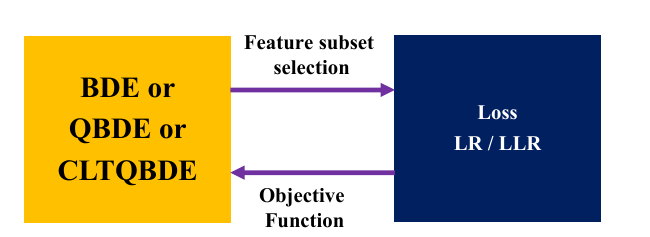}
\caption{Block diagram of the proposed wrapper}
\label{block}
\end{subfigure}
\caption{Generic framework of the proposed wrappers}
\label{fig:combined}
\end{figure*}
\section{Background and Literature Review}\label{sec2}

\subsection{Differential Evolution}
Binary Differential Evolution (BDE), a stochastic population-based global optimization algorithm, starts by initializing the random population, consisting of $N$ candidate solution vectors ($X_i$), where $N$ is the population size. This candidate solution vector follows the binary encoding scheme. Each candidate solution vector is subjected to all of the following three heuristics in each iteration (or generation) of the algorithm (see Fig. \ref{BDE}). 

At each generation $t$, the candidate solution vector ($X_i^t$) within $n$ dimensional search space, is subjected to the mutation operation yielding the mutant vector ($M_i^t$). The mutation operation is applied as presented in Eq. \eqref{eq:1}.
\begin{equation}
\label{eq:1}
M_i^t=X_{i1}^t+F* ( X_{i2}^t-X_{i3}^t )
\end{equation}
where $X_{i1}^t$,$X_{i2}^t$  and $X_{i3}^t$ are three randomly chosen distinct vectors from the current generation $t$. $F$, the mutation factor, is a user-defined parameter, and lies in the range $[0,1]$. After this, the mutant vector may not be binary anymore. Hence, sigmoid based discretization process (see Eq. \eqref{eq:2}) is applied to every $m_{ij}^t$ ($j^{th}$  member of the $M_i^t$) thereby converting continuous vector into a binary vector. 
\begin{equation}
    \label{eq:2}
   m_{ij}^t = \begin{cases}
    1, & \text{if } \text{rand}(0,1) < \text{sigmoid}(m_j^t) \\
    0, & \text{else}
\end{cases}
\end{equation}
Then, the discretized mutant vector is subjected to crossover operation where it is subjected to the mating with the corresponding candidate solution vector to generate the trial vector. The crossover operation is applied to trial vector $U_i^t$,  as presented in Eq. \eqref{eq:3}.
\begin{equation}
    \label{eq:3}
u_{ij}^t = \begin{cases}
    m_{ij}^t, & \text{if } \text{rand}(0,1) < CR \text{ and } j \neq \text{$randi$} \\
    x_{ij}^t, & \text{if } \text{rand}(0,1) \geq CR \text{ and } j \neq \text{$randi$}
\end{cases}
\end{equation}
where $j=1,2,…n, u_{ij}^t$ is the $j^{th}$ bit of $U_i^t$, $rand(0,1)$ is the random number generated in the interval $[0,1]$ from a uniform distribution. $randi$ is a randomly chosen index to make sure that the generated trial vector is different from the mutant vector. CR represents the crossover rate, is a user-defined parameter, and lies in the range $[0,1]$. 

Finally, the fitness score is computed for the trial vectors. Then, the selection operation is applied by comparing the corresponding target vectors and trial vector to produce an offspring. Better solutions survive and forms the parent population for the subsequent iteration. The selection operation follows the rule as presented in Eq.  \eqref{eq:4}:
\begin{equation}
    \label{eq:4}
X_i^{(t+1)} = \begin{cases}
    X_i^{(t)}, & \text{if } f(X_i) > f(U_i) \\
    U_i^{(t)}, & \text{otherwise}
\end{cases}
\end{equation}

As mentioned earlier, this is continued till the completion of maximum iterations or other convergence criteria, if any, are met.

\subsection{Quantum Computing}
To exploit the notions of the quantum theory within classical computers, quantum-inspired algorithms are proposed by \citep{ref1}. It employs quantum mechanics concepts such as quantum measurement, superposition of states, inference, and entanglement. A quantum bit is the basic unit of information in quantum computation defined by the linear combination of the two states as given in Eq. \eqref{quantum_state1}.
\begin{equation}
\label{quantum_state1}
Q = \alpha \ket{0} + \beta \ket{1} 
\end{equation}
The coefficients of $\alpha$ and $\beta$ are two complex numbers that must satisfy the norm relation as given in Eq. \eqref{quantum_state2}.
\begin{equation}
\label{quantum_state2}
| \alpha |^2 + | \beta |^2 = 1 
\end{equation}
where the probability of the observing state $\ket{0}$ is $|\alpha|^2$ and the probability of the state $\ket{1}$ is $|\beta|^2$. A quantum register is composed of $n$ qubits containing $2^n$ possible values simultaneously owing to the superposition of states. 

\subsection{Overview of Chaos Theory}

The theory of chaos originated in the 1800s and was further developed by \citep{ref52} to tackle challenges in complex non-linear systems \citep{ref51}. Chaotic systems are dynamic and deterministic, evolving from initial conditions, with trajectories describing the system states in the state space. Chaos heavily relies on the initial conditions and exhibits two properties: ergodicity and intrinsically stochastic nature. Chaotic maps generate the sequence of numbers that exhibit these chaotic properties, aiding EAs in escaping local minima \citep{ref48,ref49,ref50}. It is noticed that these chaotic maps are introduced to dynamically adjust the hyperparameters, and enhance adaptability to handle evolutionary dynamics optimally. It also facilitates searching in the regions which are left out by the random sequence. They enable exploration in regions left out by random sequences and have been extensively studied in large-scale continuous optimization problems. Chaotic maps produce a series of numbers from a probability distribution that differs from a uniform (0,1) distribution, exhibiting deterministic randomness. This deterministic nature allows predicting the sequence of numbers generated, as they are governed by differential equations and subject to the initial conditions. Now, a well-known chaotic map logistic map which is used in the current study is discussed below:

\textbf{Logistic map} \citep{ref28}: The Logistic map exhibits chaotic behaviour in a discrete-time demographic model. This is a polynomial mapping of degree 2. The mathematical representation is defined in Eq. \eqref{chaotic_state}.
\begin{equation}
\label{chaotic_state}
d_{t+1}  =\lambda*( d_t*( 1-d_t  ))
\end{equation}
Here, the constant $\lambda$ lies in the range of $[0,4]$ and determines the behaviour of this Logistic map. In the current research study, $\lambda$ = 4 is chosen. 



\subsection{Literature Review}
The first quantum-inspired evolutionary algorithm (QEA) was proposed by Han and Kim for the knapsack problem in the pioneering paper, \citep{ref1}. Their approach employed qubit notation and a rotation gate together with the migration operation, strategically guiding the solution to reach a global optimum. \cite{ref2} proposed an adaptive quantum differential evolution (AQDE), which dynamically adjusts the mutation and cross-over rates based on their success streaks. It outperformed QEA in solving the knapsack problem. \cite{ref3} proposed elitism-based quantum differential evolution for FSS on small datasets. \cite{ref4} proposed Multi-strategy quantum differential evolution, which integrates multiple strategies, including a mutation strategy with difference vector, multi-population mutation, and adaptive rotation angle state. Other quantum evolutionary algorithms studied in the literature include (i) hybrid of QDE (QDE acronym is not defined anywhere) and grey wolf optimizer \citep{ref5} tailored for the knapsack problem, (ii) multi-objective quantum-inspired hybrid DE \citep{ref6}, which is a hybrid of genetic algorithm quantum variants and DE for multi-objective next-release problems, and (iii) vector hop algorithm \citep{ref7}, a hybrid of differential evolution and particle swarm optimization. \cite{ref8} proposed an improved quantum differential evolution by incorporating the principles of the divide-and-conquer concept of a cooperative coevolutionary algorithm, which improved both exploration and exploitation capabilities. (iv) Another hybrid DE where the first stage invoked QDE and the resultant population is passed on to BDE in the second stage, which continues the evolution process. All these hybrids improved the search process and obtained better convergence capabilities than their individual constituents in the standalone mode.

We now briefly survey the parallel and distributed versions of DE \citep{ref12,ref13,ref14,ref15,ref16} developed across varied environments like Spark, CUDA, MPI, and OpenMP. \cite{ref15} proposed two master-slave-based parallel approaches, namely, (i) a data-based MapReduce model and (ii) a population-based MapReduce model. \cite{ref16} introduced two parallel strategies, master-slave and island approaches, evaluated on the AWS Spark cloud and tested the performance on benchmark functions. \cite{ref17} and \cite{ref18} developed master-slave-based parallel DE approaches for large-scale clustering and cluster optimization problems, respectively. \cite{ref19} introduced a cost-sensitive DE classifier (SCDE) based on Euclidean distance for imbalanced classification datasets. \cite{ref20} introduced a fine-grained parallel DE under the OpenMP framework for optimal networking, aiming to reduce the computational load on mappers and reducers. \cite{ref21} introduced Parallel DE (PDE) under the Spark environment, showing promising speedup.  \cite{ref22} presented SgtDE, a grouping topology model for large-scale optimization, achieving significant speedup. \cite{ref23} developed parallel DE under CUDA, while \citep{ref24} designed a self-adaptive DE framework in CUDA for benchmark functions. Further, several parallel versions of DE are proposed to solve resource allocation problems \citep{ref25,ref26,ref27}, hydro scheduling \citep{ref30}, large-scale clustering \citep{ref33}, optimized workflow placement \citep{ref32}, and multi-objective flow scheduling problems \citep{ref34} as well. Table A1 of the  Appendix captures a brief overview of the literature. 

\begin{figure*}[!ht]
\centering
\includegraphics[width=2\columnwidth]{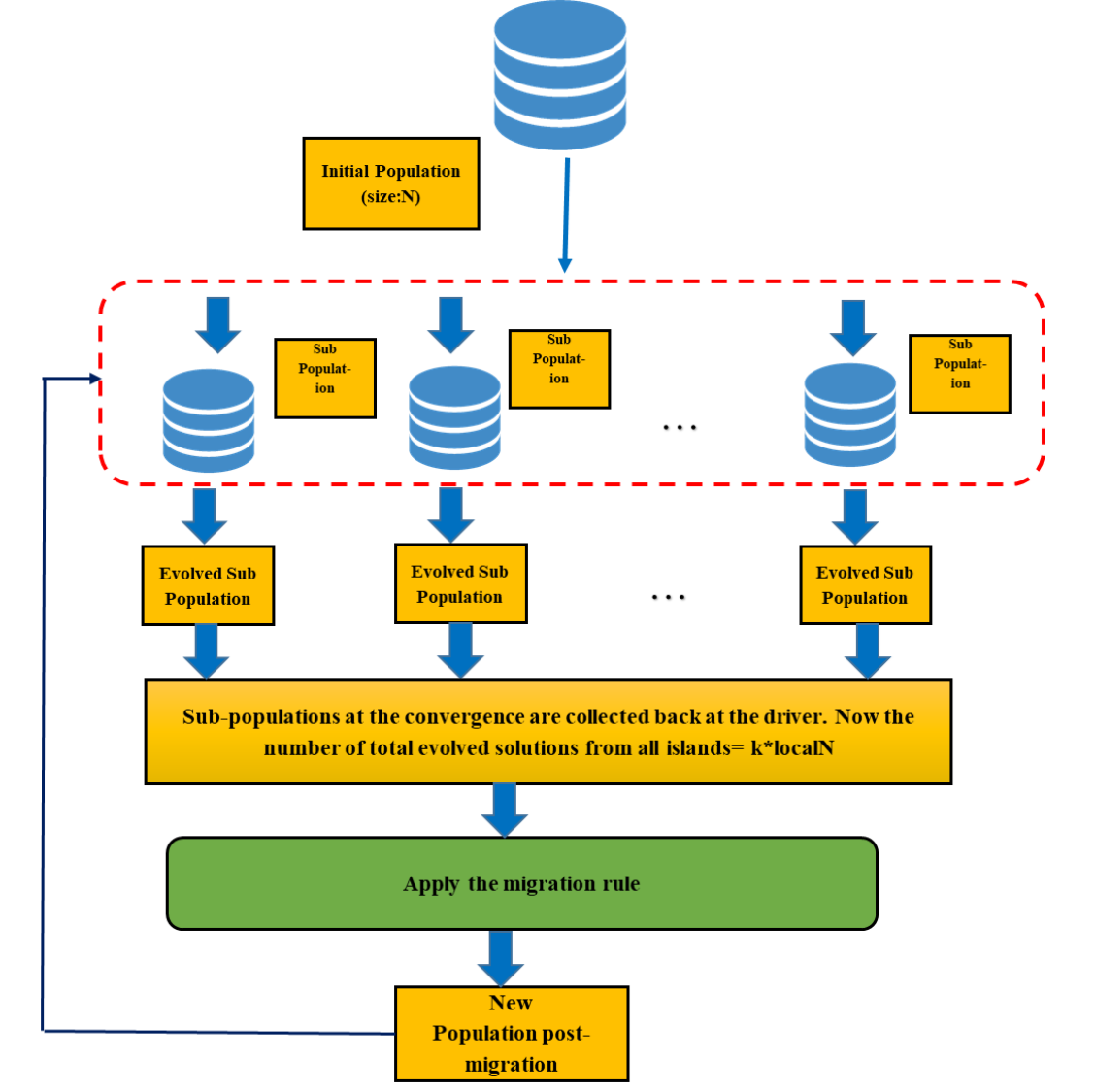}
\caption{Generic schematic diagram of the island model based wrapper}
\label{flowchart_main}
\end{figure*}
\section{Proposed Methodology}
This section introduces the objective function employed in the current study, followed by the algorithm, and an overview of the proposed parallel mechanism, which applies to all algorithms discussed. All the acronyms employed in the study are presented in Table \ref{tab:notation}.

\subsection{Objective Function}

The objective function considered in this study is the Area under the receiver operating characteristic Curve (AUC) Eq.\eqref{eq:auc}. It is defined as the mean of specificity and sensitivity. We specifically considered AUC due to its proven robustness while handling imbalanced datasets. 
\begin{equation}
         \label{eq:auc}
        \text{AUC} =  \frac{\text{Sensitivity}+\text{Specificity}}{2}
\end{equation}
Where sensitivity (refer to Eq. \eqref{eq:spec}) is the ratio of the positive samples that are correctly predicted to be positive to all the predicted positive samples. This is also called the True Positive Rate (TPR).
\begin{equation}
         \label{eq:sens}
\text{Sensitivity} = \frac{TP}{TP + FN}
\end{equation}
Where $TP$ and $FN$ are the true positive and false negative, respectively.
Specificity (refer to Eq. \eqref{eq:spec}) is the ratio of the negative samples that are correctly predicted to be negative to all the predicted negative samples. This is also called True Negative Rate (TNR).
\begin{equation}
         \label{eq:spec}
\text{Specificity} = \frac{TN}{TN + FP} 
\end{equation}
Where $TN$ and $FP$ are the true negative, and false positive, respectively.

This study employs LR and LLR as the classifiers for the proposed FSS wrapper because their training converges quickly. Unlike LR,  LLR performs regularization and reduces the number of features too.
\subsection{Proposed Chaotic Quantum Inspired Algorithm}

In this study, we propose two distinct algorithms based on chaotic and quantum principles: (i) non-gate variant, one focusing on information exchange across different qubits without gates (CLQBDE-I), and (ii) gate variant, the second involving information exchange via gates (CLQBDE-II) (see Table \ref{tab:algorithms} and Fig. \ref{QBDE}, Fig. \ref{CQBDE}).

Unlike the chaos-based algorithms proposed in the literature \cite{ref38,ref39,ref40}, the Lyapunov exponent guides our population initialization method, ensuring that the wrapper based FSS operates within a true chaotic regime. Consequently, chaotic numbers are not utilized from the initial time step of the logistic chaotic map, unlike the existing approaches. Instead, they are introduced after a specific number of time steps, which are determined by the Lyapunov exponent. After thorough experimentation, we discovered that the logistic map enters a truly chaotic regime after $5000$ time steps. This is observed offline and employed within the algorithm. Histogram plots comparing the distributions of numbers generated from the uniform random distribution and logistic chaotic map guided by Lyapunov exponent from 1 to 5000 time steps are presented in Fig. A1 of the Appendix\textit{\textbf{. }} It is crucial to note that in the current study, Lyapunov exponent-guided logistic map is introduced as the sequence generated by chaotic numbers is observed to be in true chaotic regime after first $5000$ timesteps. Consequently, both variants, CQBDE-I and CQBDE-II, incorporate chaotic numbers in the initialization for each qubit's alpha and beta states for $i^{th}$ solution, satisfying Eq.~\eqref{quantum_matrix}.
\begin{equation}
\label{quantum_matrix}
Q^{(i)} = \begin{bmatrix}
    \alpha^{(i)}_1 & \alpha^{(i)}_2 & \dots & \alpha^{(i)}_d \\
    \beta^{(i)}_1  & \beta^{(i)}_2  & \dots & \beta^{(i)}_d
\end{bmatrix}
\end{equation}
\begin{equation}
\label{condition}
\left|\alpha^{(i)}_j\right|^2 + \left|\beta^{(i)}_j\right|^2 = 1
\end{equation}
where $n$ is the number of dimensions, $i$ is the index of the solution in the population of size $N$. 

The quantum representation of each population member is given by Eq. \eqref{quantum_matrix} where the components satisfy the condition in Eq. \eqref{condition}. It consists of $N$ candidate solution vectors ($X_i$), where $n$ is the population size and each solution vector is of size $2*n$, where $n$ is the number of features. Often thus, formed quantum state vectors need to be converted into binary encoded solutions for combinatorial optimization problems. In the literature, the following rule is generally adopted. 
\begin{equation}
\label{quantum_conversion}
    x^{(i)}_{j} =
\begin{cases}
    1, & \text{if } \text{rand}(0,1) < \left|\beta^{(i)}_{j}\right|^2 \\
    0, & \text{otherwise}
\end{cases}
\end{equation}
where $i$ and $j$ are the solution and feature indices, respectively.

However, we observed that as the number of dimensions ($n$) increases, more features are being selected. This observation prompted us to introduce a constraint-based quantum state-to-binary solution conversion. The modified rule is defined as follows (refer to Eq. \eqref{quantum_conversion_theta}). This Equation is called as the threshold trick. This is employed for both the QBDE-I and QBDE-II variants.
\begin{equation}
\label{quantum_conversion_theta}
    x_i =
\begin{cases}
    1, & \text{if } \text{rand}(0,1) < \left|\beta^{(i)}_{j}\right|^2 \text{ and } d < \theta \\
    0, & \text{otherwise}
\end{cases}
\end{equation}
Here, $\theta$ represents a user-defined parameter within the range of $[0,1]$, $d$ is the number generated from rand(0,1).

We proposed two different variants of chaotic quantum algorithms guided by the Lyapunov exponent and threshold trick. The distinction between them is as follows: (i) CLQBDE-I is a non-gate variant, and (ii) CLQBDE-II is a gate-variant where a rotation gate is employed in place of quantum mutation. 

\subsubsection{CLQBDE-I}
The proposed Chaotic Quantum Binary Differential Evolution-I (CLQBDE-I) begins by generating the quantum matrix as described in Eq. \eqref{quantum_matrix}. For each candidate solution $i$, the $\alpha^{(i)}_j$ variables are generated from the chaotic series, while the corresponding $\beta^{(i)}_j$ variables are obtained by using the relation Eq.~\eqref{condition}. As mentioned earlier, the first step of the chaotic series is guided by the Lyapunov exponent. Once generated, the binary encoded solution is produced using the threshold trick outlined in Eq. \eqref{quantum_conversion_theta}. The fitness score is calculated by training the classifier, LR or LLR. 
\begin{equation}
\label{quantum_mutation_1}
\small
\begin{bmatrix}
    \alpha^{(i)}_j \\
    \beta^{(i)}_j 
\end{bmatrix}
=
\begin{bmatrix}
    \alpha^{(x1)}_j + F \cdot (\alpha^{(x2)}_j - \alpha^{(x3)}_j) \\
    \beta^{(x1)}_j + F \cdot (\beta^{(x2)}_j - \beta^{(x3)}_j) 
\end{bmatrix}
\end{equation}
Then, the initial population undergoes mutation operation (refer to Eq. \ref{quantum_mutation_1}) effected by each candidate solution's $\alpha$ and $\beta$ states by randomly selecting the three solutions. This operation is performed for a pre-specified population size. 
\begin{equation}
\label{quantum_crossover_1}
    \begin{bmatrix}
    \alpha^{(u)}_j \\
    \beta^{(u)}_j
    \end{bmatrix}
    =
    \begin{cases}
    \begin{bmatrix}
        \alpha^{(m)}_j \\
        \beta^{(m)}_j
    \end{bmatrix}, & \text{if } d < CR \text{ and } j \neq \text{$randi$} \\ \\
    \begin{bmatrix}
        \alpha^{(u)}_j \\
        \beta^{(u)}_j
    \end{bmatrix}, & \text{otherwise}
    \end{cases}
\end{equation}
where $d$ is the number generated from rand(0,1).

Subsequently, the mutated quantum matrix of each of the solutions undergoes crossover operation as described in Eq. \ref{quantum_crossover_1}. Selection is then performed by combining parent and offspring populations and sorting them based on fitness scores. The resulting population becomes the parent population for the next iteration. This iterative process continues for a pre-specified number of iterations, resulting in the convergence of the algorithm, and the resultant population is evaluated on the test data, which remains unchanged throughout the evolution process.

\subsubsection{CLQBDE-II}
We proposed another variant, gate-based quantum binary differential evolution, and named it chaotic quantum binary differential evolution-II (CQBDE-II). The distinction between them lies in employing the rotation gate in place of the quantum mutation operation to generate the mutated quantum matrix. 

The population is initialized The rotation gate is introduced to mimic the behaviour of the mutation gate. These rotation gates are represented as unitary matrices, which are employed to rotate the state of a qubit based on the rotation angle. We followed the look-up table obtained from \citep{ref1} to decide the rotation angle. The rotation gate is multiplied with each qubit of the corresponding qubit in the candidate solution and thus the mutated vector is generated. Thissolution vector is then subjected to the crossover operation, selection, and threshold trick. These steps are identical to that of CLQBDE-I as discussed in the sub-section 3.2.1. The rotation gate employed in this study is presented in Eq. \ref{eq:rotation}.
\begin{equation}
    U(\Delta \theta) =
\begin{bmatrix}
    \cos(\Delta \theta) & -\sin(\Delta \theta) \\
    \sin(\Delta \theta) & \cos(\Delta \theta)
\end{bmatrix}
\label{eq:rotation}
\end{equation}
\begin{table*}[ht]
    \centering
    \caption{Encoding Scheme of the population}
    \label{tab:encoding}
    \begin{tabular}{|l|c|}
        \hline
        \textbf{Key} & \textbf{Value:} \\
        \hline
        $Key_1$ & $\langle$ $BinaryVector_1$, $QuantumMatrix_1$, $trainedModelCoef_1$, $AUC_1$ $\rangle$ \\
        \hline
        $Key_2$ & $\langle$ $BinaryVector_2$, $QuantumMatrix_2$, $trainedModelCoef_2$, $AUC_2$ $\rangle$ \\
        \hline
        \dots & \dots \\
        \hline
        $Key_N$ & $\langle$ $BinaryVector_N$, $QuantumMatrix_N$, $trainedModelCoef_N$, $AUC_N$ $\rangle$ \\
        \hline
    \end{tabular}
\end{table*}
\subsection{Proposed Parallel Approach}
The proposed parallel algorithm adopts an island approach, which divides the data into several partitions known as data islands. 

Table \ref{tab:encoding} depicts the population schema maintained for all approaches in this work. The population consists of a solution of size N. Each solution has two different fields: (i) Key field having the unique id information to identify the solution uniquely, (ii) The Value field has the following subfields: (a) Binary vector: which is of length number of features, nfeat is a binary vector where the presence of '0' says that a particular feature is not selected, and '1' says that the particular feature is selected. (b) Quantum matrix: Here, the quantum state information of the corresponding is stored.(c) Trained model coefficients: as we know that in the wrapper methods, a classifier is chosen to evaluate the solution's performance. Thus trained model coefficients are stored in this sub-field. The main reason for storing them is to use them in the test phase. (d) AUC: The trained model results in an AUC for each solution. That information is stored in the sub-field. This kind of schema makes sure to preserve the solution-related information in a single space.

The proposed parallel approach comprises three main phases: Initialization, Training phase, and Test phase. As illustrated in Fig. \ref{flowchart_main}, there are $k$ sub-populations and $k$ data islands. The parallel algorithm operates at two levels: driver and worker. The corresponding algorithms are presented in Algorithms 1 and 2 (see Appendix), respectively. All these phases, along with migration rule invocation, happen at the driver node, while the evolution of the parallel EAs happens during the training phase, and evaluation during the test phase is executed at the worker node. However, filtering top solutions post-migration and the aggregation of the final results during the test phase is performed at the driver node.

\textbf{Phase-I: Initialization} In this phase, the quantum matrix for the population size is chaotically initialized using the biased sampling method \cite{ref10} depending on the underlying metaheuristic (i.e., DE / EA). This population follows the data structure as presented in Table \ref{tab:encoding}. Subsequently, binary-encoded solutions are obtained using a threshold trick. Thus initialized population is treated as a global population and broadcasted along with hyperparameters from the driver to the worker nodes.

\textbf{Phase-II: Training Phase} A miniature EA is evolved in parallel in each data island at each worker node. The block diagram of miniature wrapper is depicted in Fig. \ref{block}. A sub-population/local population of size $localN$ ($< N$) is initially extracted by following random sampling with replacement. The $train-and-update$ phase begins, where fitness scores are evaluated using the classifier, LR or LLR, followed by the updation of the quantum matrix, binary encoded solutions, classifier coefficients, and AUC in the respective fields (see Table \ref{tab:encoding}). This $train-and-update$ process is common for all the algorithms. Heuristics specific to each algorithm are applied to the quantum matrix of the population, resulting in the generation of an offspring quantum matrix. $Train-and-update$ operations follow, and the top $localN$ solutions are retained after selection. This process is repeated for a pre-specified maximum number of generations ($mGen$), resulting in $k * lps$ solutions from each of the $k$ data islands. Upon completing these steps, the control returns to the driver, where the migration policy is invoked.

\textbf{Migration Policy} Solutions are sorted based on the fitness scores, and the top $N$ solutions are selected. After invoking the migration policy, the worker algorithm is again executed using the population obtained post-migration which is distributed to worker nodes. Subsequently, the sub-population corresponding to a single island is selected by following random sampling with replacement and the evolution process continues. This process is repeated for a pre-specified number of migrations ($mMig$).

\textbf{Phase-III: Test Phase} In this phase, the converged population obtained in the training phase is evaluated on the test dataset. That means, AUC is computed on the test dataset using the coefficients corresponding to each solution. Subsequently, test fitness scores are computed, and the results are reported accordingly. Notably, while the proposed parallel metaheuristics differ in their respective heuristics, they adhere uniformly to the same train-and-update step, migration policy invocation, and test phase.

\section{Results \& Discussions}
All experiments are conducted on a 5-node cluster configuration, comprising a driver node and 4 slave nodes, each equipped with Intel i7 $9^{th}$ generation processors and 32GB of RAM. Notably, the driver node serves the dual role as the driver node as well as the worker node. The benchmark datasets analyzed in this study are briefly described in Table \ref{tab:dataset_info}, while Table \ref{tab:hyperparameters} presents the optimal hyperparameters identified through meticulous fine-tuning following grid-search. A train-test split ratio of 80\%:20\% is followed using stratified random sampling to ensure equal proportion of representation of classes in both datasets. All algorithms are executed for 20 runs in order to mitigate the impact of random seed variation on the algorithm, which is the standard practice followed in the evolutionary algorithms (EAs) literature. For all the experiments, the number of migrations is fixed at one. The top solution in each run is identified as the one achieving the highest AUC. Thus, we get 20 top solutions, one each for 20 runs. Mean AUC and mean cardinality of these 20 top solutions are computed and presented in  Tables  \ref{tab:bde_res}-\ref{tab:cqiea_res}.

The proposed different quantum algorithms are compared against a non-quantum counterpart, BDE, and the Chaotic Quantum-Inspired Evolution Algorithm (CQIEA)\cite{ref35}. To assess the effectiveness of the proposed method, we developed the variants listed in Table \ref{tab:my_label}.

The reasons for developing these variants are as follows: The algorithms QBDE-I and QBDE-II are meant to see the effectiveness of introducing quantum operators; CQBDE-I and CQBDE-II are designed to see the effectiveness of chaos; CLQBDE-I and CQBDE-II are designed to see the role of the Lyapunov exponent when coupled with chaotic maps. Further, CTQIEA and CLTQIEA are introduced to see the effectiveness of the threshold trick and threshold trick in combination with the Lyapunov exponent, respectively.
\begin{table*}[htbp]
    \centering
    \caption{Dataset Information}
    \label{tab:dataset_info}
    \begin{tabular}{|l|c|c|c|c|}
        \hline
        Dataset Name & \# Objects & \# Features & Size of dataset \\
        \hline
        Epsilon & 500,000 & 2,000 & 10.8 GB \\
        \hline
        IEEE Malware & 1,500,000 & 1,000 & 3.2 GB \\
        \hline
        OVA\_Omentum & 1,584 & 10,935 & 108.3 MB \\
        \hline
        OVA\_Uterus & 1,584 & 10,935 & 108.3 MB \\
        \hline
    \end{tabular}
\end{table*}
\begin{table*}[htbp]
    \centering
    \caption{Hyper parameters employed in the current study}
    \label{tab:hyperparameters}
    \begin{tabular}{|l|c|c|c|c|c|c|}
        \hline
        Dataset Name & $N$ ($localN$) & Total generations $mGen$ per migration & CR & MR & $\theta$ \\
        \hline
        Epsilon & 30 (15) & 20 (10) & 0.90 & 0.80 & 0.10 \\
        \hline
        IEEE Malware & 30 (15) & 10 (5) & 0.90 & 0.80 & 0.15 \\
        \hline
        OVA\_Omentum & 300 (200) & 20 (10) & 0.90 & 0.8 & 0.01 \\
        \hline
        OVA\_Uterus & 300 (200) & 20 (10) & 0.90 & 0.8 & 0.01 \\
        \hline
    \end{tabular}
\end{table*}
\begin{table*}[htbp]
    \centering
    \caption{Results of BDE variants}
    \label{tab:bde_res}
    \begin{tabular}{|l|c|c|c|c|c|c|c|c|}
        \hline
         &  \multicolumn{8}{c|}{Datasets}\\ 
         \hline
        Algorithm & \multicolumn{2}{c|}{Epsilon}  & \multicolumn{2}{c|}{IEEE Malware}  & \multicolumn{2}{c|}{OVA\_Omentum}  & \multicolumn{2}{c|}{OVA\_Uterus} \\
        \hline
         & $f_1$ & $f_2$ & $f_1$ & $f_2$ & $ f_1$ & $f_2$ & $f_1$ & $f_2$  \\
        \hline
        BDE+LR & 1321.3 & 0.835 & 646.50 & 0.802 & 876.26 & 0.836 & 63.35 & 0.788 \\
        \hline
        BDE+LLR & 1323.7 & 0.846 & 505.7 & 0.817 & 112.65 & 0.833 & 108.65 & 0.802 \\
        \hline
    \end{tabular}
\end{table*}
\begin{table*}[htbp]
    \centering
    \caption{Results of QBDE variants}
    \label{tab:qbde_res}
    \begin{tabular}{|l|c|c|c|c|c|c|c|c|}
        \hline
         &  \multicolumn{8}{c|}{Datasets}\\ 
         \hline
        Algorithm & \multicolumn{2}{c|}{Epsilon}  & \multicolumn{2}{c|}{IEEE Malware}  & \multicolumn{2}{c|}{OVA\_Omentum}  & \multicolumn{2}{c|}{OVA\_Uterus} \\
        \hline
         & $f_1$ & $f_2$ & $f_1$ & $f_2$ & $ f_1$ & $f_2$ & $f_1$ & $f_2$  \\
        \hline
        QBDE-I+LR & 204.7 & 0.742 & 121.25 & 0.753 & 159.25 & 0.800 & 69.90 & 0.807 \\
        \hline
        QBDE-II+LR & 214.75 & 0.745 & 127.30 & 0.758 & 70.2 & 0.846 & 70.65 & 0.808 \\
        \hline
        QBDE-I+LLR & 248.95 & 0.753 & \textbf{168.5} & \textbf{0.826} & 168.5 & 0.826 & 107.75 & 0.796 \\
        \hline
        QBDE-II+LLR & 140.85 & 0.783 & 140.6 & 0.793 & 107.0 & 0.944 & 108.4 & 0.799\\
        \hline
    \end{tabular}
\end{table*}
\begin{table*}[htbp]
    \centering
    \caption{Results of CQBDE variants}
    \label{tab:cqbde_res}
    \begin{tabular}{|l|c|c|c|c|c|c|c|c|}
        \hline
         &  \multicolumn{8}{c|}{Datasets}\\ 
         \hline
        Algorithm & \multicolumn{2}{c|}{Epsilon}  & \multicolumn{2}{c|}{IEEE Malware}  & \multicolumn{2}{c|}{OVA\_Omentum}  & \multicolumn{2}{c|}{OVA\_Uterus} \\
        \hline
         & $f_1$ & $f_2$ & $f_1$ & $f_2$ & $ f_1$ & $f_2$ & $f_1$ & $f_2$  \\
        \hline
        CQBDE-I+LR & 269.6 & 0.522 & 142.05 & 0.80 & 794.7 & 0.5 & 830.05 & 0.5 \\
        \hline
        CQBDE-II+LR & 143.85 & 0.780 & 144.5 & 0.791 & 107.5 & 0.905 & 106.45 & 0.842 \\
        \hline
        CQBDE-I+LLR & 397.95 & 0.797 & 162.55 & 0.818 & 159.25 & 0.808 & 108.75 & 0.796 \\
        \hline
        CQBDE-II+LLR & 138.65 & 0.784 & 145.85 & 0.799 & 110.35 & 0.855 & 101.7 & 0.808\\
        \hline
    \end{tabular}
\end{table*}
\begin{table*}[htbp]
    \centering
    \caption{Results of CLQBDE variants}\label{tab:clqbde_res}
    \begin{tabular}{|l|c|c|c|c|c|c|c|c|}
        \hline
         &  \multicolumn{8}{c|}{Datasets}\\ 
         \hline
        Algorithm & \multicolumn{2}{c|}{Epsilon}  & \multicolumn{2}{c|}{IEEE Malware}  & \multicolumn{2}{c|}{OVA\_Omentum}  & \multicolumn{2}{c|}{OVA\_Uterus} \\
        \hline
         & $f_1$ & $f_2$ & $f_1$ & $f_2$ & $ f_1$ & $f_2$ & $f_1$ & $f_2$  \\
        \hline
        CLQBDE-I+LR & 273.85 & 0.523 & 143.8 & 0.806 & 798.7 & 0.5 & 798.7 & 0.5 \\
        \hline
        CLQBDE-II+LR & 141.5 & 0.777 & 140.6 & 0.793 & 107 & 0.944 & 107.15 & 0.853 \\
        \hline
        CLQBDE-I+LLR & 145.95 & 0.755 & 145.95 & 0.796 & 156.95 & 0.923 & \textbf{151.0} & \textbf{0.865} \\
        \hline 
        CLQBDE-II+LLR & 141.70 & 0.784 & 140.9 & 0.801 & \textbf{110.25} & \textbf{0.954} & 112.2 & 0.812 \\
        \hline
    \end{tabular}
\end{table*}
\begin{table*}[htbp]
    \centering
    \caption{Results of CQIEA variants}
    \label{tab:cqiea_res}
    \begin{tabular}{|l|c|c|c|c|c|c|c|c|}
        \hline
         &  \multicolumn{8}{c|}{Datasets}\\ 
         \hline
        Algorithm & \multicolumn{2}{c|}{Epsilon}  & \multicolumn{2}{c|}{IEEE Malware}  & \multicolumn{2}{c|}{OVA\_Omentum}  & \multicolumn{2}{c|}{OVA\_Uterus} \\
        \hline
         & $f_1$ & $f_2$ & $f_1$ & $f_2$ & $ f_1$ & $f_2$ & $f_1$ & $f_2$  \\
        \hline
        CQIEA+LR & 1968.10 & 0.841 & 885.5 & 0.802 & 5033.0 & 0.788 & 6942.9 & 0.792 \\
        \hline
        CTQIEA+LR & 421.0 & 0.790 & 217.65 & 0.779 & 240.6 & 0.947 & 241.1 & 0.884 \\
        \hline
        CLTQIEA+LR & 412.7 & 0.790 & 208.3 & 0.812 & 290.0 & 0.819 & 280.95 & 0.794 \\
        \hline
        CQIEA+LLR & 888.0 & 0.853 & 677.9 & 0.810 & 8356.25 & 0.837 & 6942.90 & 0.792 \\
        \hline
        CTQIEA+LLR & \textbf{256.20} & \textbf{0.815} & 190.8 & 0.798 & 261.50 & 0.842 & 318.8 & 0.806 \\
        \hline
        CLTQIEA+LLR & 258.50 & 0.815 & 192.05 & 0.801 & 258.70 & 0.893 & 332.3 & 0.803 \\
        \hline
    \end{tabular}
\end{table*}
\begin{table*}[htbp]
    \centering
    \caption{Results of the paired t-test}
    \label{tab:ttest_results}
    \begin{tabular}{|l|c|c|c|c|c|}
        \hline
        Dataset & Model (Top1 vs Top2 w.r.t AUC) & t-statistic & p-value \\
        \hline
        Epsilon & \textbf{CTQIEA + LLR\text{*}} vs  & 0.421 & \textbf{0.675} \\
        &  CLTQIEA+LLR & &  \\
        
        \hline
        IEEE Malware & \textbf{QBDE-I+LLR\text{*}} vs  & 1.204 & \textbf{0.235} \\
        & CQBDE-I + LLR & &  \\
        \hline
        OVA\_Omentum & \textbf{CLQBDE-II + LLR\text{*}} vs  & 1.14 & \textbf{0.260} \\
        &  CTQIEA + LR & &  \\
        \hline
        OVA\_Uterus & CTQIEA + LR vs  & 3.84 & 0.0004 \\
        & \textbf{CLQBDE-I + LLR\text{**}} & & \\
        \hline
        \multicolumn{4}{l}{\small *Better method based on cost-benefit analysis.} \\
        \multicolumn{4}{l}{\small **Better method based on  statistical significance.}
     \end{tabular}
\end{table*}

\subsection{Comparative Analysis Ablation Study}
We conducted our ablation study in a boosted manner. That means in the first instance, parallel BDE is developed as a baseline and then quantum-inspired versions of BDE are developed (boosted wrapper-I), and finally, chaotic versions of these quantum-inspired algorithms are developed (boosted wrapper-II).

Initially, non-quantum variants, i.e., BDE variants, are developed as a baseline, and the corresponding results are presented in Table \ref{tab:bde_res}. With an aim to remove irrelevant features, we employed LLR in place of LR as a classifier. LLR demonstrated its ability to weed out unimportant features in the following datasets: (i) In the IEEE Malware dataset, BDE with LLR as the classifier, obtained mean AUC, and there was a reduction of 150 features. (ii) Similarly, in the OVA\_Omentum dataset, LLR led to a great reduction (around 600 features) in terms of mean cardinality, with slightly improved mean AUC. However, this observation is not consistently noticed across the remaining two datasets. In the other datasets (Epsilon and OVA\_Uterus), the mean AUC is slightly improved. However, it is accompanied by a modest increase in the mean cardinality. 

The strength of LLR over LR is that the latter retains unimportant features during the evolution process, leading to increased cardinality of the solutions.

In the boosted wrapper-I, to obtain better optimal results, we developed their quantum-inspired counterparts, namely,  QBDE-I and QBDE-II, where the QBDE-I variant has gates, and the QBDE-II variant has no gates. Consequently, we achieved lower mean cardinality accompanied by higher mean AUC across all datasets except the Epsilon dataset (see Table \ref{tab:qbde_res}). 

Then, in the boosted wrapper-II, to further improve the exploration capability, we introduced chaotic initialization in two different ways: i.e., with and without Lyapunov-guided chaotic series (see Table \ref{tab:cqbde_res}, (see Table \ref{tab:clqbde_res})) resulting in CLQBDE and CQBDE respectively. In the latter case (i.e., without the Lyapunov exponent), we noticed a reduction in mean AUC in all the datasets than their corresponding non-chaotic counterparts (i.e. QBDE variants). This is likely due to the fact that we do not operate in a true chaotic regime.

To balance both the exploration and exploitation capability of an algorithm, we introduced Lyapunov-guided chaotic series (see Table ~\ref{tab:clqbde_res})in the initialization phase. It turned out that these variants achieved efficient solutions meaning higher mean AUC with decreased mean cardinality. The chaotic numbers introduced in the first step (of CQBDE variants) are not generated in a truly chaotic regime. However, after introducing the Lyapunov exponent-guided chaotic numbers (i.e. CLQBDE variants), we discarded the first 5000 time steps, ensuring the chaotic series is in the true chaotic regime. This transition helped the algorithm improve exploration and exploitation capability as evidenced by the results in Table \ref{tab:clqbde_res}). This is spectacularly observed in the case of both high-dimensional datasets, i.e., OVA\_Omentum and OVA\_Uterus. However, the true chaotic regime alone did not suffice in the other two datasets. 

Further, to create a level-playing field to the algorithm we would like to compare with (i.e. variants of CQIEA \citep{ref35} and CLQIEA), we demonstrated the effectiveness of the threshold trick and the influence of Lyapunov exponent-guided chaotic numbers by invoking them in the variants of CQIEA \citep{ref35} and CLQIEA in Table \ref{tab:cqiea_res}. We noticed that the original CQIEA \citep{ref35}, i.e., without the threshold trick and the Lyapunov exponent-guided chaotic numbers, yielded unacceptably high mean cardinality, impacting its performance adversely. However, after introducing the threshold trick combined with Lyapunov exponent-guided chaotic numbers, we could effectively control the cardinality while improving the mean AUC. This behaviour is particularly evident when LASSO LR (LLR) is employed as the classifier. 

\subsection{Statistical Testing}
A two-tailed t-test at 5\%  level of significance and 38 (=20+20-2) degrees of freedom is conducted on the mean AUC obtained from 20 runs across the top two algorithms (w.r.t mean AUC) in each dataset (refer to Table \ref{tab:ttest_results}). The t-test demonstrates that in three out of four datasets (except OVA\_Uterus), the top 2 best algorithms solely based on AUC,  turned out to be statistically similar. The top 2 best algorithms corresponding to each dataset are as follows: (i) in the Epsilon dataset (CTQIEA+LLR and CLTQIEA+LLR), (ii) in IEEE Malware dataset (QBDE-I+LLR and CQBDE-I+LLR), (iii) in OVA\_Omentum dataset (CLQBDE-II+LLR and CTQIEA+LR) and (iv) in OVA\_Uterus dataset (CTQIEA+LR and CLQBDE-I+LLR).

In the case of statistical similarity, to break the tie, preference is accorded to the algorithm that selected less mean cardinality. In other words, higher preference is accorded to the algorithm that yielded a great reduction in mean cardinality and an insignificant reduction in mean AUC. After performing this type of cost-benefit analysis, the better-performing algorithm is presented in bold face in Table \ref{tab:ttest_results}).

It is noticed that in three datasets, the top-performing algorithm yielded a higher mean AUC with almost similar mean cardinality. In these cases, preference is accorded to the one with a higher mean AUC. For example, in the IEEE Malware dataset, QBDE-I+LLR obtained a mean AUC of 0.826 with a mean cardinality of 168.5. However, the next-best algorithm (CQBDE-I+LLR) obtained a mean AUC of 0.818 (which is $<$ 0.008 than that obtained by QBDE-I+LLR) with a mean cardinality of 162.55. Here, the reduction in cardinality is very minimal ($<$ 6 features). This makes the QBDE-I+LLR win over the CQBDE-I algorithm. The same cost-benefit analysis is followed for the other datasets as well, resulting in the Epsilon and OVA\_Omentum datasets; the QBDE-I + LLR and CLQBDE-II+LLR are the winners, respectively. However, in the OVA\_Uterus dataset, CLQBDE-I + LLR turned out to be statistically significant when compared to the CTQIEA-I + LR. Accordingly, CLQBDE-I + LLR is presented in boldface in Table \ref{tab:ttest_results}

It is important to note that when algorithm A and algorithm B are statistically similar, the winner is chosen based on the cost-benefit analysis. However, if statistical significance is observed, the preference is accorded to the algorithm, which is the statistically significant algorithm. 

In summary, the insights derived from the current study are as follows:
\begin{itemize}
    \item In the Epsilon dataset, CTQIEA + LLR emerged as the winner due to the significant impact of the threshold trick and chaos.
    \item In the IEEE Malware dataset, QBDE-I+LLR was the best algorithm, primarily due to the effectiveness of the threshold trick.
    \item In the OVA\_Omentum dataset, CLQBDE-II + LLR outperformed others due to the introduction of Lyapunov-based chaos and the threshold trick.
    \item In the OVA\_Uterus dataset, CLQBDE-I + LLR was the winner largely due to the guidance by Lyapunov.
    \item The CLQBDE variants with LLR as a classifier turned out to be the best wrappers no matter a quantum gate is adopted or not in both high-dimensional datasets.
\end{itemize}

\section{Conclusions}
This study proposes CLQBDE-I, where a Lyapunov exponent-guided chaotic map-based initialization method is incorporated into the quantum-inspired BDE algorithm for FSS. Our results show its superiority on high-dimensional datasets and its competitive performance on Epsilon and IEEE Malware datasets compared to the alternative algorithms. CLQBDE-I outperformed not only our baseline, namely, QBDE but also other baselines CLTQIEA-I and CTQIEA-I in all but the IEEE Malware dataset in the latter algorithm. Overall, working in conjunction with QBDE variants, integrating Lyapunov exponent-guided chaotic dynamics into them yielded better solutions than simple, chaotic-based initialization methods. 

Future research directions include extending this approach to multi-objective environments, designing hybrid chaotic mapping techniques, and exploring chaotic quantum hybrid EAs. This methodology also holds promise for large-scale clustering and feature selection applications across various domains like finance and economics. 

\backmatter

\bibliography{sn-article}

\begin{thebibliography}{}
\providecommand{\doi}[1]{\url{https://doi.org/#1}}
\bibcommenthead

\bibitem[\protect\citeauthoryear{Adhianto, Banerjee, Fagan, Krentel, Marin, Mellor-Crummey, and Tallent}{Adhianto et~al.}{2020}]{ref20}
Adhianto, L., S.~Banerjee, M.~Fagan, M.~Krentel, G.~Marin, J.~Mellor-Crummey, and N.~Tallent. 2020.
\newblock Hpctoolkit: Tools for performance analysis of optimized parallel programs.
\newblock {\em Concurr. Comput. Pract. Exp.\/}~22: 685--701 .

\bibitem[\protect\citeauthoryear{Al-Sawwa and Ludwig}{Al-Sawwa and Ludwig}{2020}]{ref19}
Al-Sawwa, J. and S.~Ludwig. 2020.
\newblock Performance evaluation of a cost-sensitive differential evolution classifier using spark -- imbalanced binary classification.
\newblock {\em J. Comput. Sci.\/}~40: 101065 .

\bibitem[\protect\citeauthoryear{Cao, Zhao, Lv, and Liu}{Cao et~al.}{2017}]{ref25}
Cao, B., J.~Zhao, Z.~Lv, and X.~Liu. 2017.
\newblock A distributed parallel cooperative coevolutionary multiobjective evolutionary algorithm for large-scale optimization.
\newblock {\em IEEE Transactions on Industrial Informatics\/}~13: 2030--2038 .

\bibitem[\protect\citeauthoryear{Chandrashekar and Sahin}{Chandrashekar and Sahin}{2014}]{ref39}
Chandrashekar, B. and F.~Sahin. 2014.
\newblock A survey on feature selection methods.
\newblock {\em Comput. Electr. Eng.\/}~40: 16--28 .

\bibitem[\protect\citeauthoryear{Chen, Jiang, Li, Li, and Wang}{Chen et~al.}{2016}]{ref18}
Chen, Z., X.~Jiang, J.~Li, S.~Li, and L.~Wang. 2016.
\newblock Pdeco: Parallel differential evolution for clusters optimization.
\newblock {\em J. Comput. Chem.\/}~34: 1046--1059 .

\bibitem[\protect\citeauthoryear{Cho, Nyunt, and Aung}{Cho et~al.}{2019}]{ref17}
Cho, P., T.~Nyunt, and T.~Aung 2019.
\newblock Differential evolution for large-scale clustering.
\newblock In {\em Proc. 2019 9th Int. Work. Comput. Sci. Eng. (WCSE 2019 SPRING)}, pp.\  58--62.

\bibitem[\protect\citeauthoryear{Danforth}{Danforth}{2013}]{ref52}
Danforth, C.M. 2013.
\newblock Chaos in an atmosphere hanging on a wall.
\newblock {\em Mathematics of Planet Earth 2013\/} .

\bibitem[\protect\citeauthoryear{Das, Mullick, and Suganthan}{Das et~al.}{2016}]{ref41}
Das, S., S.S. Mullick, and P.N. Suganthan. 2016.
\newblock Recent advances in differential evolution--an updated survey.
\newblock {\em Swarm and Evolutionary Computation\/}~27: 1--30 .

\bibitem[\protect\citeauthoryear{Das and Suganthan}{Das and Suganthan}{2011}]{ref0}
Das, S. and P.~Suganthan. 2011, Feb.
\newblock Differential evolution: A survey of the state-of-the-art.
\newblock {\em IEEE Transactions on Evolutionary Computation\/}~{\em 15\/}(1): 4--31 .

\bibitem[\protect\citeauthoryear{de~P.~Veronese and Krohling}{de~P.~Veronese and Krohling}{2010}]{ref23}
de~P.~Veronese, L. and R.~Krohling 2010.
\newblock Differential evolution algorithm on the gpu with c-cuda.
\newblock In {\em IEEE Congress on Evolutionary Computation}, pp.\  1--7.

\bibitem[\protect\citeauthoryear{Deng, Tan, Dong, and Tan}{Deng et~al.}{2015}]{ref21}
Deng, C., X.~Tan, X.~Dong, and Y.~Tan. 2015.
\newblock A parallel version of differential evolution based on resilient distributed datasets model, {\em Commun. Comput. Inf. Sci.}, Volume 562,  84--93.

\bibitem[\protect\citeauthoryear{Deng, Shang, Cai, Zhao, Zhou, Chen, and Deng}{Deng et~al.}{2021}]{ref8}
Deng, W., S.~Shang, X.~Cai, H.~Zhao, Y.~Zhou, H.~Chen, and W.~Deng. 2021.
\newblock Quantum differential evolution with cooperative coevolution framework and hybrid mutation strategy for large scale optimization.
\newblock {\em Knowledge-Based Systems\/}~224: 107080.
\newblock \doi{10.1016/j.knosys.2021.107080} .

\bibitem[\protect\citeauthoryear{Deng, Xu, Gao, and Zhao}{Deng et~al.}{2022}]{ref4}
Deng, W., J.~Xu, X.Z. Gao, and H.~Zhao. 2022, Mar.
\newblock An enhanced msiqde algorithm with novel multiple strategies for global optimization problems.
\newblock {\em IEEE Transactions on Systems, Man, and Cybernetics: Systems\/}~{\em 52\/}(3): 1578--1587.
\newblock \doi{10.1109/TSMC.2020.3030792} .

\bibitem[\protect\citeauthoryear{Falco, Scafuri, Tarantino, and Cioppa}{Falco et~al.}{2017}]{ref27}
Falco, I.D., U.~Scafuri, E.~Tarantino, and A.D. Cioppa 2017.
\newblock A distributed differential evolution approach for mapping in a grid environment.
\newblock In {\em 15th EUROMICRO International Conference on Parallel, Distributed and Network-Based Processing (PDP'07)}, pp.\  442--449.

\bibitem[\protect\citeauthoryear{Ge, Yu, Lin, Gong, Zhan, Chen, and Zhang}{Ge et~al.}{2018}]{ref26}
Ge, Y., W.~Yu, Y.~Lin, Y.~Gong, Z.~Zhan, W.~Chen, and J.~Zhang. 2018.
\newblock Distributed differential evolution based on adaptive mergence and split for large-scale optimization.
\newblock {\em IEEE Transactions on Cybernetics\/}~48: 2166--2180 .

\bibitem[\protect\citeauthoryear{Glotic, Kitak, Pihler, and Ticar}{Glotic et~al.}{2014}]{ref30}
Glotic, A., P.~Kitak, J.~Pihler, and I.~Ticar. 2014.
\newblock Parallel self-adaptive differential evolution algorithm for solving short-term hydro scheduling problem.
\newblock {\em IEEE Transactions on Power Systems\/}~29: 2347--2358 .

\bibitem[\protect\citeauthoryear{Gupta, Sharma, and Jindal}{Gupta et~al.}{2016}]{ref45}
Gupta, P., A.~Sharma, and R.~Jindal. 2016.
\newblock Scalable machine-learning algorithms for big data analytics: A comprehensive review.
\newblock {\em Wiley Interdisciplinary Reviews: Data Mining and Knowledge Discovery\/}~{\em 6\/}(6): 194--214 .

\bibitem[\protect\citeauthoryear{Han, Wang, Tang, Weng, Li, and Dobre}{Han et~al.}{2021}]{ref7}
Han, D., J.~Wang, C.~Tang, T.~Weng, K.~Li, and C.~Dobre. 2021.
\newblock A multi-objective distance vector-hop localization algorithm based on differential evolution quantum particle swarm optimization.
\newblock {\em International Journal of Communication Systems\/}~{\em 34\/}(14): e4924 .

\bibitem[\protect\citeauthoryear{Han and Kim}{Han and Kim}{2002}]{ref1}
Han, K.H. and J.H. Kim. 2002, Dec.
\newblock Quantum-inspired evolutionary algorithm for a class of combinatorial optimization.
\newblock {\em IEEE Transactions on Evolutionary Computation\/}~{\em 6\/}(6): 580--593.
\newblock \doi{10.1109/TEVC.2002.804320} .

\bibitem[\protect\citeauthoryear{Harada, Kaidan, and Thawonmas}{Harada et~al.}{2020}]{ref12}
Harada, T., M.~Kaidan, and R.~Thawonmas. 2020.
\newblock Comparison of synchronous and asynchronous parallelization of extreme surrogate-assisted multi-objective evolutionary algorithm.
\newblock {\em Natural Computing\/} .

\bibitem[\protect\citeauthoryear{He, Peng, Chen, Deng, and Wu}{He et~al.}{2021}]{ref22}
He, Z., H.~Peng, J.~Chen, C.~Deng, and Z.~Wu. 2021.
\newblock A spark-based differential evolution with grouping topology model for large-scale global optimization.
\newblock {\em Cluster Comput.\/}~24: 515--535 .

\bibitem[\protect\citeauthoryear{Hota and Pat}{Hota and Pat}{2010}]{ref2}
Hota, A.R. and A.~Pat 2010.
\newblock An adaptive quantum-inspired differential evolution algorithm for 0-1 knapsack problem.
\newblock In {\em 2010 Second World Congress on Nature and Biologically Inspired Computing (NaBIC)}, Kitakyushu, Japan, pp.\  703--708.

\bibitem[\protect\citeauthoryear{Kromer, Platos, and Snasel}{Kromer et~al.}{2013}]{ref33}
Kromer, P., J.~Platos, and V.~Snasel 2013.
\newblock Scalable differential evolution for many-core and clusters in unified parallel c.
\newblock In {\em 2013 IEEE International Conference on Cybernetics (CYBCO)}, pp.\  180--185.

\bibitem[\protect\citeauthoryear{Kumari, Srinivas, and Gupta}{Kumari et~al.}{2013}]{ref6}
Kumari, A.C., K.~Srinivas, and M.~Gupta. 2013.
\newblock Software requirements optimization using multi-objective quantum-inspired hybrid differential evolution, In {\em EVOLVE - A Bridge between Probability, Set Oriented Numerics, and Evolutionary Computation II},  ed. et~al., O.S., Volume 175 of {\em Advances in Intelligent Systems and Computing}. Springer, Berlin, Heidelberg.
\newblock \doi{10.1007/978-3-642-31519-0\_7}.

\bibitem[\protect\citeauthoryear{Liu, Gao, and Wang}{Liu et~al.}{2015}]{ref49}
Liu, T., X.~Gao, and L.~Wang. 2015.
\newblock Multi-objective optimization method using an improved nsga-ii algorithm for oil–gas production process.
\newblock {\em Journal of the Taiwan Institute of Chemical Engineers\/}~57: 42--53 .

\bibitem[\protect\citeauthoryear{Lu, Niu, Liu, and Zhu}{Lu et~al.}{2013}]{ref50}
Lu, H., R.~Niu, J.~Liu, and Z.~Zhu. 2013.
\newblock A chaotic non-dominated sorting genetic algorithm for the multi-objective automatic test task scheduling problem.
\newblock {\em Applied Soft Computing\/}~{\em 13\/}(5): 2790--2802 .

\bibitem[\protect\citeauthoryear{Maier, Razavi, Kapelan, Matott, Kasprzyk, and Tolson}{Maier et~al.}{2019}]{ref43}
Maier, H.R., S.~Razavi, Z.~Kapelan, L.S. Matott, J.~Kasprzyk, and B.A. Tolson. 2019.
\newblock Introductory overview: Optimization using evolutionary algorithms and other metaheuristics.
\newblock {\em Environmental Modelling \& Software\/}~114: 195--213 .

\bibitem[\protect\citeauthoryear{May}{May}{1976}]{ref28}
May, R. 1976.
\newblock Simple mathematical models with very complicated dynamics.
\newblock {\em Nature\/}~261: 459--467 .

\bibitem[\protect\citeauthoryear{Olyaei, Wu, and Kinsner}{Olyaei et~al.}{2017}]{ref38}
Olyaei, A., C.~Wu, and W.~Kinsner. 2017.
\newblock Detecting unstable periodic orbits in chaotic time series using synchronization.
\newblock {\em American Physical Society\/}~96 .

\bibitem[\protect\citeauthoryear{Packard, Crutchfield, Farmer, and Shaw}{Packard et~al.}{1980}]{ref51}
Packard, N.H., J.P. Crutchfield, J.D. Farmer, and R.S. Shaw. 1980.
\newblock Geometry from a time series.
\newblock {\em Phys. Rev. Lett.\/}~45: 712 .

\bibitem[\protect\citeauthoryear{Pan and Da}{Pan and Da}{2015}]{ref48}
Pan, I. and S.~Da. 2015.
\newblock Fractional-order load-frequency control of interconnected power systems using chaotic multi-objective optimization.
\newblock {\em Applied Soft Computing\/}~29: 328--344 .

\bibitem[\protect\citeauthoryear{Pant, Zaheer, Garcia-Hernandez, and Abraham}{Pant et~al.}{2020}]{ref42}
Pant, M., H.~Zaheer, L.~Garcia-Hernandez, and A.~Abraham. 2020.
\newblock Differential evolution: A review of more than two decades of research.
\newblock {\em Engineering Applications of Artificial Intelligence\/}~90: 103479 .

\bibitem[\protect\citeauthoryear{Peralta, Río, Ramírez-Gallego, Triguero, Benitez, and Herrera}{Peralta et~al.}{2015}]{ref13}
Peralta, D., S.D. Río, S.~Ramírez-Gallego, I.~Triguero, J.~Benitez, and F.~Herrera. 2015.
\newblock Evolutionary feature selection for big data classification: A mapreduce approach.
\newblock {\em Math. Probl. Eng.\/} .

\bibitem[\protect\citeauthoryear{Qian, Wang, Huang, Wang, and Wang}{Qian et~al.}{2009}]{ref34}
Qian, B., L.~Wang, D.~Huang, W.~Wang, and X.~Wang. 2009.
\newblock An effective hybrid de-based algorithm for multi-objective flow shop scheduling with limited buffers.
\newblock {\em Computers \& Operations Research\/}~{\em 36\/}(1): 209--233 .

\bibitem[\protect\citeauthoryear{Ramos and Vellasco}{Ramos and Vellasco}{2020}]{ref35}
Ramos, A.C. and M.~Vellasco 2020.
\newblock Chaotic quantum-inspired evolutionary algorithm: enhancing feature selection in bci.
\newblock In {\em 2020 IEEE Congress on Evolutionary Computation (CEC)}, Glasgow, UK, pp.\  1--8.

\bibitem[\protect\citeauthoryear{Rastogi and Shim}{Rastogi and Shim}{1999}]{ref46}
Rastogi, R. and K.~Shim 1999.
\newblock Scalable algorithms for mining large databases.
\newblock In {\em Tutorial Notes of the Fifth ACM SIGKDD International Conference on Knowledge Discovery and Data Mining}.

\bibitem[\protect\citeauthoryear{Rong, Gong, and Gao}{Rong et~al.}{2019}]{ref14}
Rong, M., D.~Gong, and X.~Gao. 2019.
\newblock Feature selection and its use in big data: Challenges, methods, and trends.
\newblock {\em IEEE Access\/}~7: 19709--19725 .

\bibitem[\protect\citeauthoryear{Schliemann, Khaetskii, and Loss}{Schliemann et~al.}{2002}]{ref01}
Schliemann, J., A.V. Khaetskii, and D.~Loss. 2002.
\newblock Spin decay and quantum parallelism.
\newblock {\em Physical Review B\/}~{\em 66\/}(24): 245303 .

\bibitem[\protect\citeauthoryear{Srikrishna, Ghosh, Ravi, and Deb}{Srikrishna et~al.}{2015}]{ref3}
Srikrishna, V., R.~Ghosh, V.~Ravi, and K.~Deb. 2015.
\newblock Elitist quantum-inspired differential evolution based wrapper for feature subset selection, In {\em Multi-disciplinary Trends in Artificial Intelligence. MIWAI 2015},  eds. Bikakis, A. and X.~Zheng, Volume 9426 of {\em Lecture Notes in Computer Science}. Springer, Cham.
\newblock \doi{10.1007/978-3-319-26181-2\_11}.

\bibitem[\protect\citeauthoryear{Teijeiro, Pardo, González, Banga, and Doallo}{Teijeiro et~al.}{2016}]{ref16}
Teijeiro, D., X.~Pardo, P.~González, J.~Banga, and R.~Doallo. 2016.
\newblock Implementing parallel differential evolution on spark, In {\em Applications of Evolutionary Computation. EvoApplications 2016},  eds. Squillero, G. and P.~Burelli, Volume 9598 of {\em Lecture Notes in Computer Science}. Springer, Cham.

\bibitem[\protect\citeauthoryear{Thomert, Bhattacharya, Caron, Gadireddy, and Lefevre}{Thomert et~al.}{2016}]{ref32}
Thomert, D., A.~Bhattacharya, E.~Caron, K.~Gadireddy, and L.~Lefevre 2016.
\newblock Parallel differential evolution approach for cloud workflow placements under simultaneous optimization of multiple objectives.
\newblock In {\em 2016 IEEE Congress on Evolutionary Computation (CEC)}, pp.\  822--829.

\bibitem[\protect\citeauthoryear{Vivek, Ravi, and RadhaKrishna}{Vivek et~al.}{2022}]{ref10}
Vivek, Y., V.~Ravi, and P.~RadhaKrishna. 2022.
\newblock Scalable feature subset selection for big data using parallel hybrid evolutionary algorithm based wrapper under apache spark environment.
\newblock {\em Cluster Computing\/}.
\newblock \doi{10.1007/s10586-022-03725-w} .

\bibitem[\protect\citeauthoryear{Wang and Wang}{Wang and Wang}{2021}]{ref5}
Wang, Y. and W.~Wang. 2021.
\newblock Quantum-inspired differential evolution with grey wolf optimizer for 0-1 knapsack problem.
\newblock {\em Mathematics\/}~{\em 9\/}(1233).
\newblock \doi{10.3390/math9111233} .

\bibitem[\protect\citeauthoryear{Wong, Qin, Wang, and Shi}{Wong et~al.}{2015}]{ref24}
Wong, T., A.~Qin, S.~Wang, and Y.~Shi. 2015.
\newblock cusade: A cuda-based parallel self-adaptive differential evolution algorithm.
\newblock {\em IEEE Congress on Evolutionary Computation (CEC)\/}~2: 375--388 .

\bibitem[\protect\citeauthoryear{Wu, Mallipeddi, and Suganthan}{Wu et~al.}{2019}]{ref44}
Wu, G., R.~Mallipeddi, and P.N. Suganthan. 2019.
\newblock Ensemble strategies for population-based optimization algorithms--a survey.
\newblock {\em Swarm and Evolutionary Computation\/}~44: 695--711 .

\bibitem[\protect\citeauthoryear{Xue, Zhang, Browne, and Yao}{Xue et~al.}{2016}]{ref40}
Xue, B., M.~Zhang, W.N. Browne, and X.~Yao. 2016.
\newblock A survey on evolutionary computation approaches to feature selection.
\newblock {\em IEEE Trans. Evol. Comput.\/}~20: 606--626 .

\bibitem[\protect\citeauthoryear{Zaharia, Chowdhury, Franklin, Shenker, and Stoica}{Zaharia et~al.}{2010}]{ref47}
Zaharia, M., M.~Chowdhury, M.J. Franklin, S.~Shenker, and I.~Stoica 2010.
\newblock Spark: Cluster computing with working sets.
\newblock In {\em 2nd USENIX Workshop on Hot Topics in Cloud Computing (HotCloud 10)}.

\bibitem[\protect\citeauthoryear{Zhou}{Zhou}{2010}]{ref15}
Zhou, C. 2010.
\newblock Fast parallelization of differential evolution algorithm using mapreduce.
\newblock In {\em Proc. 12th Annu. Genet. Evol. Comput. Conf. GECCO '10}, pp.\  1113--1114.

\end{thebibliography}


\clearpage

\begin{appendices}
\section{Details of the proposed approach}
\begin{figure*}[hbt]
\centering
\includegraphics[width=1.65\columnwidth]{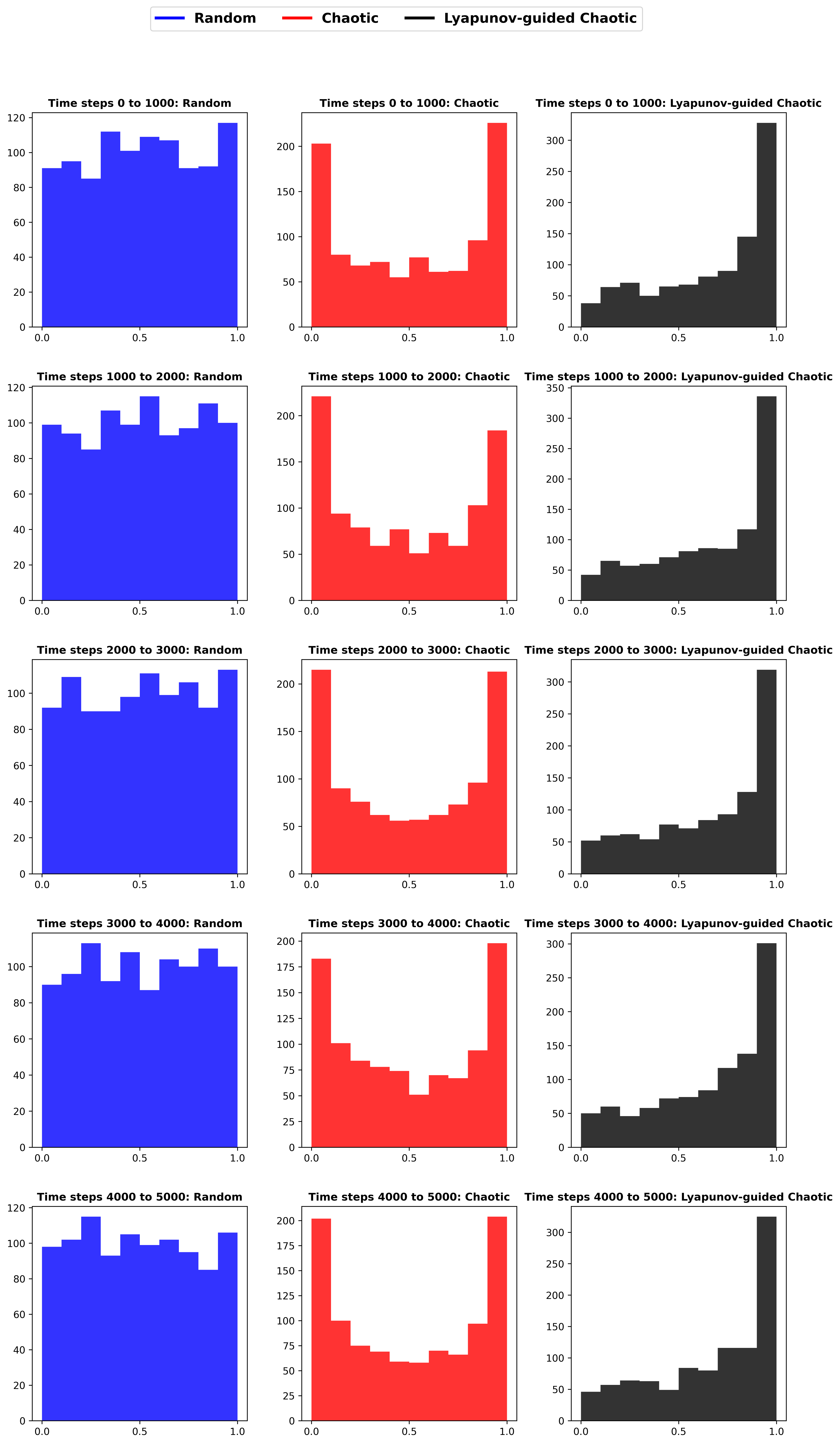}
\caption{Histograms}
\label{t100}
\end{figure*}
\begin{table*}[hbtp]
    \centering
    \caption{Summary of parallel evolutionary algorithms in different environments}\label{tab:algorithm_summary}
    \begin{tabular}{|p{2.5cm}|p{2.5cm}|p{2.5cm}|p{4cm}|}
        \hline
        Authors & Algorithm & Environment & Solved Problem \\
        \hline
        Zhou [17] & DE	& Spark	& Discussed Pros and cons of various parallel approaches \\ \hline 
        Teijeiro et al. [18] & DE	& Spark + AWS	& Benchmark functions \\ \hline 
        Cho et al. [19]	& DE & 	Spark	&  Clustering \\ \hline
        Chen et al. [20] & Modified DE & 	SPMD & 	Cluster Optimization \\ \hline
        Al-Sawwa and 
        Ludwig [21]	&  DE	&  Spark & 	DE based classifier \\ \hline
        Adhianto et al. [22] & 	DE & 	OpenMP	&  Optical Network problem \\ \hline
        Deng et al. [23]	&  DE	& Spark	& Benchmark functions  \\ \hline
        He et al. [24]	& Five variants of DE & 	Spark + Cloud	& Ring topology model applied to  benchmark functions \\ \hline
        Veronse \& Krohling [25] & 	DE	& CUDA & Large scale optimization  \\ \hline
        Wong et al. [26]	& Self-Adaptive DE	&  CUDA & Benchmark functions \\ \hline
        Cao et al. [27]	& DPCCMOEA	&  MPI	&  Large scale optimization \\ \hline
        Ge et al. [28] & DDE-AMS & 	MPI	& Large scale optimization \\ \hline
        Falco et al. [29]	& DE	& MPI	& Resource allocation \\ \hline
        Glotic et al. [30]	& PSADE	& MATLAB	& Hydro Scheduling algorithm \\ \hline
        Daoudi et al. [31]	& DE	& Hadoop	& Clustering \\ \hline
        Thomert et al. [32]	& NSDE-II	& OpenMP	& Cloud work placement \\ \hline
        Kromer et al. [33]	& DE	& Unified Parallel C	& Large scale optimization \\ \hline
        Qian et al. [34]	& MPFPSP	& Multithreading	& Flow scheduling problem \\ \hline
        Vivek et al. [12]	& PB-TADE, PB-DETA,PB-DE &	Spark &	FSS \\ \hline
        \textbf{Current study} & \textbf{CLTQBDE-I, CLTQIEA} & \textbf{Spark} & \textbf{FSS} \\
        \hline
    \end{tabular}
\end{table*}
\begin{algorithm}[hbt]
    \caption{Driver Algorithm}
    \label{algo:algorithm_name}
    \begin{algorithmic}[1]
        \State \textbf{Input:} $ps$, $lps$, $X$, $Xtrain$, $Xtest$, $mMig$, $mGen$
        \State \textbf{Output:} $P$: population evolved after $mMig$ migrations
        
        \State $i \leftarrow 0$
        \State $Qt \leftarrow$ Chaotically Initialize the quantum matrix
        \State $X \leftarrow$ getBinarySolution($Qt$)
        \State $P \leftarrow [\phi,\phi,\ldots,\phi]_{ps \times 4}$
        
        \For{$i = 0,1,\ldots, ps$}
            \State $sol \leftarrow [i,X[i], \phi, 0.0]$ \Comment{Encoding Scheme (see Table \ref{tab:encoding})}
            \State $P \leftarrow P \cup sol$
        \EndFor
        
        \State $(Xtrain_1,Xtrain_2,\ldots,Xtrain_k) \leftarrow Xtrain$ \Comment{Divide data into $k$ islands}
        
        \While{$i = 0,1,\ldots, mMig$}
            \State Call IslandMapper($P$) \Comment{Migration Rule}
            \State $R \leftarrow$ Collect $\{p_{r_1},p_{r_2},\ldots,p_{r_k}\}$ from $k$ islands
            \State $R \leftarrow$ SortBasedOnFitness($R$)
            \State $P \leftarrow \{R : R \text{ until } |R| < ps\}$
        \EndWhile
        
        \For{each sol in $P_r$}
            \State $model \leftarrow$ sol[2] \Comment{Collect coefficients for each solution}
            \State $score \leftarrow$ testModel($model, Xtest$) \Comment{Evaluate AUC on test dataset}
            \State $cardinality \leftarrow$ sum(sol[1])
            \State $testFitness \leftarrow (score \times (1 - (cardinality/m)))$
        \EndFor
        
        \State \textbf{Return} $P$, $testFitness$
    \end{algorithmic}
\end{algorithm}
\begin{algorithm}[hbt]
\caption{Worker algorithm}
\begin{algorithmic}[1]

    \State \textbf{Input:} $lps$, $mGen$, $F$, $CR$
    \State \textbf{Output:} $localP$: population evolved after $mGen$ maximum iterations

    \State $k \gets 0$
    \State localP $\gets$ Randomly pick lps number of solutions from P \Comment{local population}
    \For{$i = 0, 1, \ldots, \text{lps}$}
        \State bVC $\gets$ localP[i].BinaryEncodedVector
        \State auc $\gets$ LLR (bVC, $islandData$)
        \State localP[i][3] $\gets$ updateAUC
    \EndFor
    \For{$k = 0, 1, \ldots, \text{mGen}$}
        \State mV $\gets \{\phi\}$
        \For{$i = 0, 1, \ldots, \text{lps}$}
            \State loPVec $\gets$ localP[i].QuantumMatrix
            \State mVec $\gets$ Mutation(loPVec)
            \State mV $\cup \{mVec\}$
        \EndFor
        \State locOfP $\gets \{\phi\}$
        \For{$i = 0, 1, \ldots, \text{lps}$}
            \State mV $\gets$ mutatedP[i].QuantumMatrix
            \State TV $\gets$ Invoke Crossover on mV
            \State locOfP $\cup \{\text{TV}\}$
        \EndFor
        \State Invoke train-and-update phase on locOfP 
        \State newP $\gets \{\phi\}$
        \For{$i = 0, 1, \ldots, \text{lps}$}
            \State nS $\gets$ Invoke Selection (locOfP[i], P[i])
            \State newP $\cup \{\text{nS}\}$
        \EndFor
        \State localP $\gets$ newP
        \State $k \gets k + 1$
    \EndFor
    \State \textbf{return} localP

\end{algorithmic}
\end{algorithm}
\end{appendices}

\end{document}